\pdfoutput=1

\documentclass[11pt]{article}

\usepackage[]{acl}

\usepackage{times}
\usepackage{latexsym}

\usepackage[T1]{fontenc}

\usepackage[utf8]{inputenc}

\usepackage{microtype}

\usepackage{inconsolata}

\usepackage{graphicx}
\usepackage{enumitem,kantlipsum}

\usepackage{booktabs}
\usepackage{hyperref}
\usepackage{tabularx}
\usepackage{longtable}
\usepackage{fontawesome5}

\usepackage{times}
\usepackage{latexsym}
\usepackage{hyperref}
\usepackage{amsmath}
\usepackage{amssymb}
\usepackage{array}
\usepackage{makecell}

\title{How do Multimodal Foundation Models Encode Text and Speech?\\
An Analysis of Cross-Lingual and Cross-Modal Representations}

\author{Hyunji Lee \phantom{\and} Danni Liu \phantom{\and}  Supriti Sinhamahapatra \phantom{\and} Jan Niehues \\
        Karlsruhe Institute of Technology, Germany \\
        \texttt{hyunji.lee@student.kit.edu}, \texttt{\{firstname.lastname\}@kit.edu}
        }

\begin{document}
\maketitle

\begin{abstract}
Multimodal foundation models aim to create a unified representation space that abstracts away from surface features like language syntax or modality differences. 
To investigate this, we study the internal representations of three recent models, 
analyzing the model activations from semantically equivalent sentences across languages in the text and speech modalities. 
Our findings reveal that:
\textbf{1)} Cross-modal representations converge over model layers, except in the initial layers specialized at text and speech processing.
\textbf{2)} Length adaptation is crucial for reducing the cross-modal gap between text and speech, although current approaches' effectiveness is primarily limited to high-resource languages.
\textbf{3)} Speech exhibits larger cross-lingual differences than text.
\textbf{4)} For models not explicitly trained for modality-agnostic representations, the modality gap is more prominent than the language gap.
\end{abstract}

\section{Introduction}
Recent progress in foundation models has sparked growing interest in expanding their text processing capabilities \cite{nllb,vicuna2023,qwen} to speech
\cite{seamless,qwen-audio,salmonn,llama3}.
Despite the empirical successes,
understandings of these models' internal representations remain limited, 
particularly on 
\textit{language differences}, 
\textit{modality gaps}, 
and the impact of \textit{model architectures}.
This work aims to fill this gap by studying how text and speech are represented in recent multimodal foundation models. 

While the internal representations of multilingual models have been extensively studied, most prior works focus on single-modality analyses of text \cite{kudugunta-etal-2019-investigating,sun-etal-2023-towards-deep} or speech \cite{DBLP:conf/nips/BelinkovG17,DBLP:conf/interspeech/SeysselLADW22,icassp23_analyzing_discrete_units,sun-etal-2023-towards-deep,DBLP:conf/icassp/KheirAC24}.
Moreover, as many multimodal foundation models have dedicated subparts for languages or modalities, not all analysis techniques are directly applicable. 
For instance, similarity retrieval tasks \cite{conneau-etal-2020-emerging,icassp23_shared_speech_text,chen-etal-2023-xsim} often require identical input feature dimensions, which is not always guaranteed for speech and text.
Probing \cite{probing,DBLP:conf/nips/BelinkovG17,DBLP:conf/interspeech/SeysselLADW22} features with different dimensions leads to auxiliary classifiers of varying sizes and may skew the results.
In this work, we use Singular Vector Canonical Correlation Analysis (SVCCA; \citeauthor{svcca}, \citeyear{svcca}) due to its invariance to affine transformations, 
which is suitable for comparing features from different architectures and dimensions that occurs frequently in speech-text representations.

\begin{figure}[t]
\includegraphics[width=\linewidth,trim={22 8 20 10},clip]{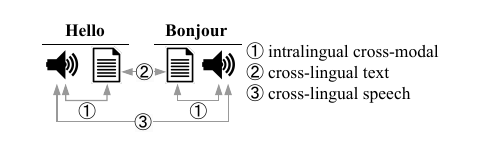}
  \caption{We use the similarity between model activations for the same sentences in different languages and modalities to measure language and modality gaps.}
  \label{fig:overview}  
\end{figure}

Previous studies comparing speech-text representations do not involve the cross-lingual aspect, 
and use either task-specific \cite{9746815,tsiamas-etal-2024-pushing} or proprietary \cite{icassp23_shared_speech_text} models.
Recent studies on multilingual representations in large language models reveal different levels of language invariance depending on training data \cite{wendler-etal-2024-llamas} and model scales \cite{lingua_franca}. 
The study most related to ours is probably the concurrent work from \citet{wu2024semantichubhypothesislanguage}, who show that representations for
semantically equivalent multilingual/modal inputs are similar in model intermediate
layers.
Our study differs in its focus on languages and modality gaps as well as its broad coverage of 30 languages at different resource levels.
To the best of our knowledge,  
we present the first cross-modal and cross-lingual analysis of representations over a wide variety of language in multimodal foundation models.

\begin{figure*}[ht!]
  \includegraphics[width=\linewidth,trim={0 7 0 7},clip] {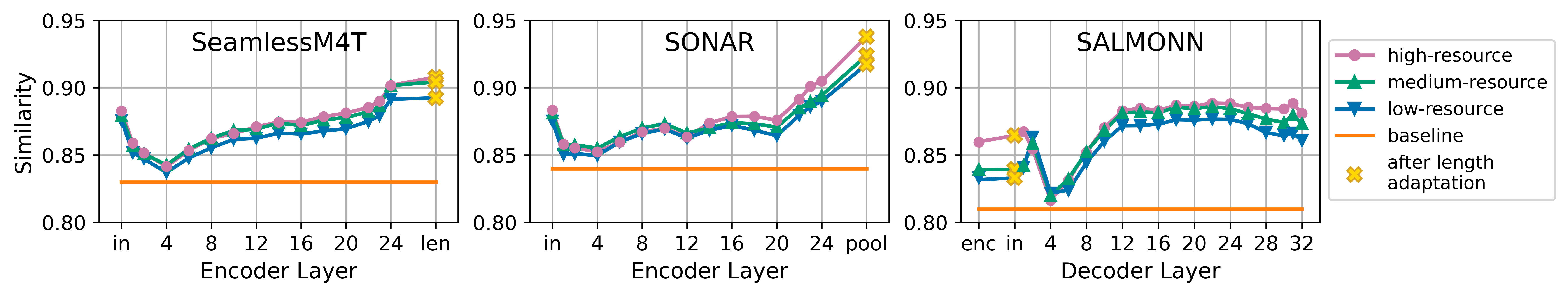}
  \caption{Average cross-modal similarity over all languages over model layers. 
  X-axis markers ``in'': input word embeddings or audio features, 
  ``len'':  after length adaptor in Seamless, ``pool'': after pooling in SONAR, ``enc'': after the frozen audio encoder, before length adaptation by window-level Q-Former in SALMONN.
  }
  \label{fig:crossmodal}
\end{figure*}

\section{Methodology}
An assumption in unified multimodal and multilingual models is that inputs are transformed into a semantic space independent of input forms.
This abstraction from surface level motivates our method.

\paragraph{Measuring similarity between semantically equivalent sentences:} 
As shown in~\autoref{fig:overview}, 
we begin with semantically equivalent sentences in different languages and modalities.
To compare their model activations at different layers, we extract these activations and employ SVCCA.
Its invariance to affine transformations \cite{svcca} ensures comparabilitiy of activations across different modalities and languages, even when they originate from different model subparts.
Given the extracted activations,
we calculate the 
SVCCA scores between speech and text versions of the same sentence (intra-lingual cross-modal) and between translations (cross-lingual text/speech).
Higher SVCCA scores indicate higher similarity.
More explanations of SVCCA scores are in~\autoref{appendix:svcca}.

\paragraph{Model selection:}
Model architectures introduce inductive biases in the learned representations.
For speech and text representations, 
a critical factor is the significant length difference between speech utterances and texts.
To explore different \textit{architectures} and, in particular, \textit{length adaptation} mechanisms,
we analyze the following models:
\begin{itemize}[nolistsep,leftmargin=*]
    \item \textbf{Seamless} \cite{seamless}: encoder-decoder model with dedicated text and speech encoders, where the latter is followed by a \textit{length adaptor} \cite{madapter} to \textit{downsample} by a fixed factor. 
    We analyze its encoder representations, as the decoder does not support parallel comparisons speech and text.
    \item \textbf{SONAR} \cite{sonar}: sentence embedding model with a multilingual text encoder and a set of monolingual speech encoders. 
    It creates \textit{fixed-size} embeddings by \textit{pooling} over sequence lengths, and is explicitly trained to align multilingual and multimodal embeddings.
    \item \textbf{SALMONN} \cite{salmonn}: decoder-only LLM (Vicuna; \citeauthor{vicuna2023}, \citeyear{vicuna2023}) adapted to ingest audio inputs. 
    It \textit{downsamples} encoded audio representations\footnote{encoded by the encoder of Whisper \cite{whisper} and the BEATS encoder \cite{beats} (both frozen)} by \textit{window-level Q-Former} \cite{blip2,salmonn} by a fixed factor. 
    We do not analyze the audio encoders' internal representations as they are audio-only.
\end{itemize}
Detailed model descriptions and our hidden representation extraction procedures are in~\autoref{appendix:model_details}.

\paragraph{Data and language:}
We use the FLEURS dataset \cite{fleurs}, 
which contains $n$-way parallel speech dataset with their transcripts from the FLoRes-101 dataset \cite{goyal-etal-2022-flores}.
We use its test split and analyze 30 languages from diverse resource levels, language families, and scripts as detailed in~\autoref{appendix:language_details}.
Due to differences in supported languages among the models, six of the 30 languages are not shared between SONAR and the others. 
To maximize comparability, we select these six languages to have the same resource level.

\paragraph{Baseline similarity:}
We calculate SVCCA scores between random vectors of the same sizes as the analyzed representations as baselines.
This represents the state of no similarity at all.

\section{Results}
We analyze 
cross-modal (\S\ref{subsec:crossmodal}) and
cross-lingual (\S\ref{subsec:crosslingual}) representations, 
and compare the impact of modality and language differences (\S\ref{subsec:crossmodal_vs_crosslingual}).

\subsection{Cross-Modal Analysis} \label{subsec:crossmodal}

\autoref{fig:crossmodal} shows the average speech-text similarity of language grouped by language resource levels.

\begin{figure*}[ht!]
  \includegraphics[width=\linewidth,trim={0 6.5 0 7},clip]{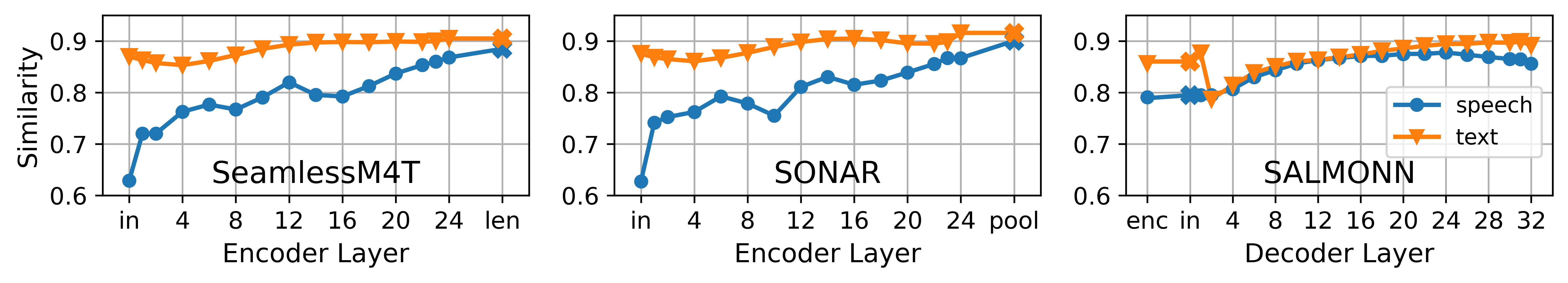}
  \caption{Average cross-lingual similarities between all language pairs in speech/text modality over model layers.}
  \label{fig:crosslingual}
\end{figure*}

\begin{figure*}[h]
  \includegraphics[width=\linewidth,trim={0 10.4 0 7.5},clip]{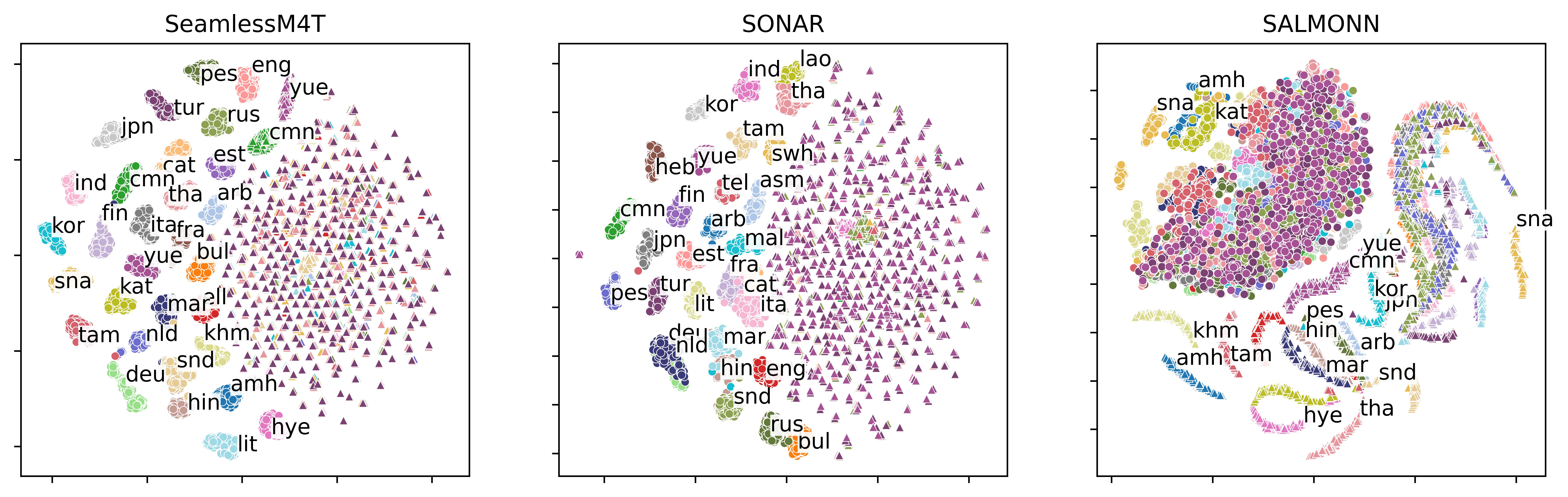}
  \caption{To visually verify how the models progressively process language and modality gaps, we use 2D visualization
  with t-SNE  \cite{JMLR:v9:vandermaaten08a} for speech and text at a middle layer (14\textsuperscript{th}, 14\textsuperscript{th}, 18\textsuperscript{th} from left to right).
  For Seamless and SONAR, texts are organized by semantics while speech remains clustered by language or language family.
  For SALMONN, languages with diverse scripts remain distinct in text representations.}
  \label{fig:tsne}
\end{figure*}

\paragraph{Progression through layers:}
Generally, cross-modal similarity increases with the number of layers, as expected. This suggests a growing abstraction of semantic meaning independent of the input modalities.
However, all three models consistently exhibit a dip in cross-modal similarity at the initial layers.
We believe this is related to the different functionalities of the earlier layers in audio and text processing models.
While the early layers of text processing models primarily capture syntactic information \cite{belinkov-etal-2017-neural,peters-etal-2018-deep},
audio encoders tend to focus on acoustic features like speaker identity in the lower layers \cite{DBLP:conf/interspeech/ChungHTG19,wavlm}.
After this initial specialized processing, 
representations for both modalities exhibit more similarity based on their semantics.
Moreover, cross-modal similarity for SALMONN flattens and drops slightly at later layers. 
This is likely due to its decoder-only architecture which generates text outputs only and makes the model diverge from the shared representations that are learned for both modalities.

\paragraph{Impact of language resource level:}
As shown in the lower part of~\autoref{fig:crossmodal},
the overall trend of increasing similarity over the layers remains consistent across all resource levels. 
However, lower-resource languages consistently exhibit lower similarity scores, 
suggesting that they are less effectively mapped into a shared representation space than their higher-resource counterparts.

\paragraph{Impact of length adaptor:}
In~\autoref{fig:crossmodal}, by comparing the yellow crosses with their preceding data points, 
we can assess the impact of the length adaptors.
SONAR’s pooling mechanism, coupled with its dedicated losses, are the most effective in minimizing the modality gap.\footnote{This complete elimination of the length difference also limits the model's expressiveness and therefore performance on downstream tasks, as shown in \citet{sonar}.}
For Seamless and SALMONN, while their length adaptor and window-level Q-Former exhibit a slight positive impact in reducing the modality gap, this effect appears limited to high- and medium-resource languages.
In low-resource languages, as evidenced by the flat slope towards the yellow crosses in ~\autoref{fig:crossmodal}, these length adaptation mechanisms do not seem to be as effective.
This limitation may be attributed to weaker representations for speech in lower-resource languages, hindering the learning of effective shrinking mechanisms.

\begin{figure*}[ht!]
\centering
  \includegraphics[width=0.98\linewidth,trim={7 10 7 7},clip]
  {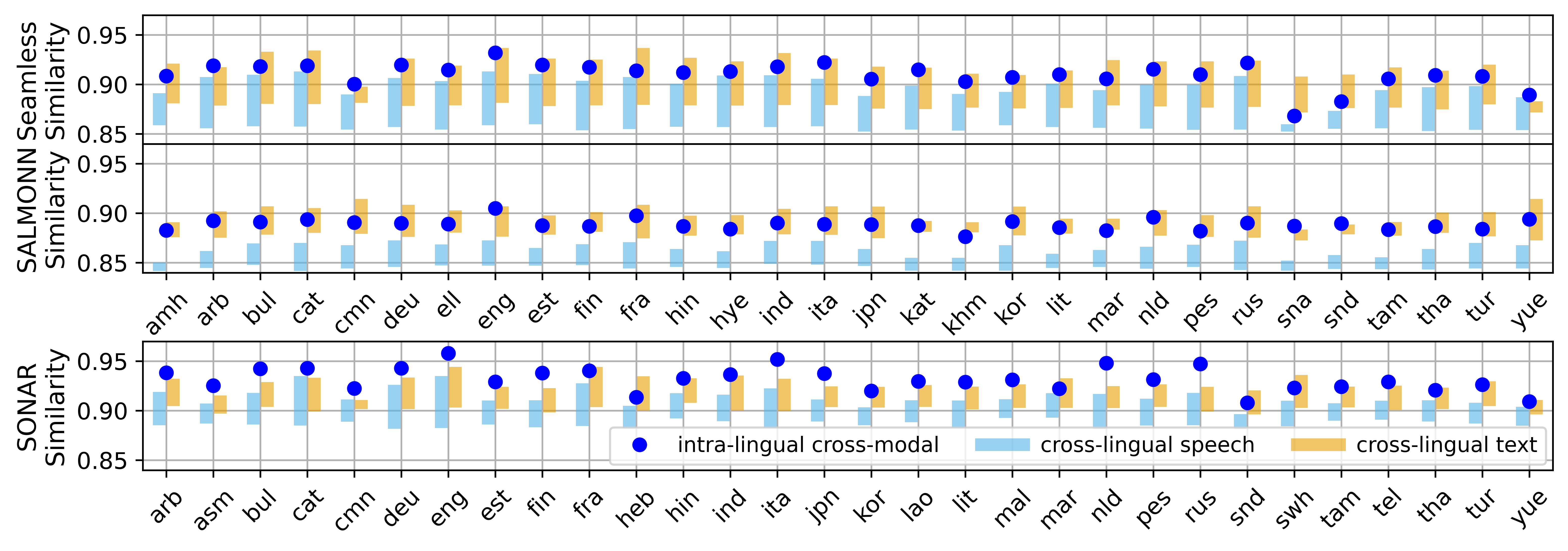}
  \caption{Given representations of text sentences at the last layer in one language, similarity to the same sentences in speech (``intra-lingual cross-modal''), their translations in text (``cross-lingual text''), and their translations in speech (``cross-lingual speech''). 
  Latter two shown as range over all 29 language pairs. Language codes in~\autoref{tab:language_statistics}.}
  \label{fig:compare_text_speech}
\end{figure*}

\subsection{Cross-Lingual Analysis} \label{subsec:crosslingual}
After assessing the representational differences across modalities (\S\ref{subsec:crossmodal}), 
we hold modality constant and examine cross-lingual differences. 

\paragraph{Higher overall cross-lingual similarity in text than speech:}
\autoref{fig:crosslingual} shows the average cross-lingual similarities for speech and text across model layers. 
Overall, with the exception of the initial layers in SALMONN, the higher cross-lingual similarities observed for text suggest that
the models more effectively create a unified cross-lingual space for the text modality compared to speech.
This is likely due to the greater variability in speech, as the same utterance can be expressed in various ways, different in vocal characteristics, speaking pace, and recording conditions. 
In contrast, text typically adheres to a single, standardized form of writing.
This  greater variability can pose more challenges in abstracting towards semantic representations independent of input languages.

\paragraph{Initial drop in SALMONN text cross-lingual similarity due to fragmented tokenization:}
We suppose that the initial drop in cross-lingual similarity in the text modality within~\autoref{fig:crosslingual} is related to insufficient tokenizer coverage for diverse languages.
As Vicuna’s vocabulary size is limited to 32$k$ (inherited from LLaMA \cite{llama}), many languages with diverse scripts are inadequately supported.
This results in texts being tokenized at the character or byte level, which are shared across many languages, inflating initial similarity in the input embeddings but settling soon in the subsequent layers.
To visually verify this hypothesis, we use t-SNE plots.
The visualizations for SALMONN in Figure \ref{fig:tsne} further support the tokenizer deficiency, as text representations that remain distinct are predominantly languages with diverse scripts, such as Khmer (khm), Armenian (hye), and Japanese (jpn).
To further support this finding, 
we quantified the relation between fragmented tokenization and similarity scores after the word embedding layer. 
We calculated the Pearson correlation coefficient between the proportaion of shared tokens between parallel sentences in two languages (averaged over all sentences in FLoRes devtest set) and their pairwise similarity scores on SALMONN.
Over the 435 language pairs, we found a positive correlation coefficient of 0.228 ($p$-value=1.48e-06).\footnote{
This relation may be even stronger after disentangling the effects of resource level.
Fragmented tokenization often occurs in lower-resource languages, which in general have lower similarity scores (as shown in \autoref{fig:crossmodal}).
}

\paragraph{Language gap reduced earlier in text than speech:}
In the other plots in~\autoref{fig:tsne} for Seamless and SONAR, the language gap appears to be reduced earlier in text than in speech. 
While text data points are primarily organized by semantics in a middle layer, speech data points are still clustered by language or language family, as evidenced by clusters by language or language families like Sindhi (snd) and Hindi (hin). 
This aligns with our previous findings on higher overall cross-lingual similarity in text compared to speech.

\subsection{Comparing Language and Modality Gaps}
\label{subsec:crossmodal_vs_crosslingual}
Our previous analyses have held language or modality constant and compared cross-modal and cross-lingual differences.
A logical next step is to study the relative influence of the modality gap and the language gap.
As shown in~\autoref{fig:compare_text_speech}, 
for Seamless and SALMONN, intra-lingual cross-modal similarity (between text and speech of the same sentence) is mostly always lower than the highest cross-lingual text similarity (between a sentence and its text translation).
This means that for texts in a given language, there exists a text translation in another, presumably related, language that is representationally more similar than the same sentences in speech. 
It also implies that for these models, the modality gap is larger than the language gap.
This is somewhat counter-intuitive since the intra-lingual cross-modal setup involves the same language, but can be explained by previous findings on modality differences such as length mismatch.

The picture is slightly different with SONAR, which was explicitly trained to bridge  modality and language gaps.
For most languages, intra-lingual cross-modal similarity surpasses other types of similarity, 
suggesting that SONAR more effectively reduces the modality gap than the language gap.
The different observations from SONAR highlight that, 
unless explicitly optimized for reduction, 
both modality and language gaps persist in multimodal foundation models, with the modality gap often being more pronounced than language gaps.

\section{Conclusion}
To study how multimodal foundation models process text and speech across diverse languages, 
we analyzed their internal representations based on the  similarity of semantically equivalent sentences.
Our findings highlight that while these models present a \textit{unified architecture} for handling various modalities and languages, 
they do not inherently create \textit{fully unified representations} by semantic meaning.
Representational gaps, some of which already observed in task-specific models, 
including speech-text length mismatches (e.g., \citealt{gaido-etal-2021-ctc,madapter}), 
weak representation for low-resource languages,
and tokenizer bottleneck (e.g., \citealt{zhang-etal-2022-robust}, \citealt{salesky-etal-2023-multilingual}),
still persist in current multimodal foundation models.

Besides the findings presented earlier, our study offers several practical recommendations.
The first is to incorporate representation analyses into the development cycle of models, especially on models designed to reduce modality gaps.
Another recommendation is model choices for speech-text downstream tasks. 
Practitioners working with low-resource or zero-shot use cases may consider initializing their models with foundation models explicitly trained for closing modality and language gaps.

\section*{Limitations}
\paragraph{Data and Modality Coverage} First, our study is limited by its reliance on multiway aligned text and speech data, which is scarce. 
Specifically, our findings are based on the FLEURS dataset \cite{fleurs}, which is created from Wikipedia texts. 
This may limit the generalizability of our findings to other domains, such as informal or spoken texts.
Additionally, this study focuses on two modalities of speech and text. 
Exploring other modalities like images would be very interesting. 
However, as our research question focuses on speech-text foundation models, 
we consider analyzing other modalities out of the scope of the current work.

\paragraph{Model Coverage} In this study, we analyzed three multimodal foundation models widely-adopted at the beginning of this project.
Since then, many new multimodal models supporting text and speech have emerged, such as
Qwen-Audio \cite{qwen-audio} and Llama 3 \cite{llama3}.
Extending our analyses to more of these models would be a valuable addition. 
Nonetheless, we believe our coverage of encoder-decoder, sentence embedding, and decoder-only architectures, including the decoder-only SALMONN model, provides a sufficiently diverse representation of model types.

\paragraph{Language Coverage} Due to differences in supported languages among the analyzed models, six of the 30 languages are not shared between SONAR and the two models. 
A fully overlapping set of languages would have provided a cleaner experimental setup. 
However, since our conclusions are based on comparisons of similarity scores within the same model between modalities and languages, rather than across different models, we believe that the differing sets of languages do not compromise the validity of our findings.

\paragraph{Analysis Type}
Our findings are only drawn from intrinsic analyses based on feature vectors, i.e., SVCCA scores on activations. 
Additional results by other explainability methods will complement the current findings, 
e.g., Logit lens \cite{logitlens} or performance on downstream tasks.

\paragraph{Prompts Variation}
We do not vary prompts in the experiments on SALMONN, meanwhile NLLB and SONAR do not support prompting. 
Prompting multimodal large language models itself is activate research field, and to the best of our knowledge, there is no established prompt for bridging speech-text modality gaps. 
Given the analysis-oriented nature of this work, we did not focus on prompt optimization.
However, it would be interesting as the field of MLLM prompting advances.

\section*{Acknowledgments}
Part of this work was performed on the HoreKa supercomputer funded by the
Ministry of Science, Research and the Arts Baden-Württemberg and by
the Federal Ministry of Education and Research.
Part of this work was supported by funding from the pilot program Core-Informatics of the Helmholtz Association (HGF).
Part of this work received support from
the European Union’s Horizon research and innovation programme under grant agreement No
101135798, project Meetween (My Personal AI Mediator for Virtual MEETtings BetWEEN People).

\bibliography{anthology_0,anthology_1,custom}

\appendix
\section{Details on SVCCA}
\label{appendix:svcca}
We use the Singular Vector Canonical Correlation Analysis (SVCCA; \citeauthor{svcca}, \citeyear{svcca}) to evaluate the similarly of the extracted speech and text representations.
Given two sets of representations $X \in \mathbb{R}^{F_x{\times}M}$ and $Y \in \mathbb{R}^{F_y{\times}N}$, 
where $M$ and $N$ are the number of data points 
and $F_x$ and $F_y$ are the dimension of the features, 
SVCCA measures their similarity.
The inputs may differ in their feature dimension ($F_x \neq F_y$), 
but the number of data points must be the same ($M$ = $N$). 
SVCCA first performs a singular value decomposition (SVD) on both $X$ and $Y$, 
resulting in two sets of singular vectors and singular values. 
Then, Canonical Correlation Analysis (CCA) is applied on only the top $m \leq M$ and top $n \leq N$ singular vectors that explain $k\%$ variance of $X$ and $Y$. 
CCA will then find linear transformations that maximize the correlation between two vector sets, returning CCA correlation coefficients. 
The averaged value of all coefficients is the SVCCA similarity value $\in [0, 1]$, depending on how similar ($=1$) or different ($=0$) the two sets of representations are.

We use the implementation from \citet{svcca}\footnote{\url{https://github.com/google/svcca}}.
We take singular vectors that explain $90\%$ variance of in the data.
To stabilize the similarity computations, we use an \texttt{epsilon} of \texttt{1e-10}. 
To compare variable-length sequences, we follow prior works \cite{kudugunta-etal-2019-investigating,liu-etal-2021-improving-zero,sun-etal-2023-towards-deep} and meanpool over the sequence length dimension.


\section{Details on Models and Hidden Representation Extraction} 
\label{appendix:model_details}

\subsection{SeamlessM4T}
Seamless \textbf{M}assively \textbf{M}ultilingual \& \textbf{M}ultimodal \textbf{M}achine \textbf{T}ranslation (SeamlessM4T; \citeauthor{seamless}, \citeyear{seamless}) is an encoder-decoder model that supports speech-to-text, text-to-text, and speech-to-speech translation/transcription. 
It covers over 100 languages.
Text inputs go through the text encoder and decoder, 
which are initialized with SeamlessM4T-NLLB \cite{seamless}, 
a multilingual text-to-text translation model supporting 200 languages. 
Speech inputs first pass through the mel filterbank feature extraction, 
where the outputs are given to the Conformer speech encoder, 
initialized with the speech representation learning model W2v-BERT 2.0 \cite{w2vBERT2021} and is followd by a length adaptor. 
The length adaptor of SeamlessM4T is a modified version of the M-Adaptor \cite{madapter}, which downsamples the speech with a fixed factor.
We focus on the encoder, as the subsequent parts do not support both text and speech in parallel.

We use \texttt{seamless-m4t-v2-large}\footnote{\url{https://huggingface.co/facebook/seamless-m4t-v2-large}} for the analyses.
Both speech and text encoders have $24$ layers with a feature size of $1024$. 
We analyze the speech and text representations after the layers $\{1, 2, 4, 6, 8, 10, 12, 14, 16, 18, 20, 22, 23, 24\}$, 
both speech and text input embeddings, and the speech representations after the length adaptor.

\subsection{SONAR}

\textbf{S}entence-level multim\textbf{O}dal and la\textbf{N}guage-\textbf{A}gnostic \textbf{R}epresentations (SONAR; \citeauthor{sonar}, \citeyear{sonar}) is a multimodal and multilingual sentence embedding model for 200 languages.
It has \textit{one multilingual} text encoder initialized with NLLB \cite{nllb} and \textit{multiple monolingual} speech encoders initialized with Wav2Vec2-BERT 2.0 
\cite{w2vBERT2021}.
A multilingual text decoder initialized with NLLB is used in training for translation and autoencoding.
The encoder outputs are used to produce sentence embeddings by pooling along the sequence length dimension (meanpooling for text and learned attention pooling is used for speech). 
Additionally, the mean squared error (MSE) loss is used on the encoder outputs, 
which encourages aligning sentences in the shared embedding space by reducing the differences between embeddings of the same semantic meaning but of different languages and modality. 

We use the pre-trained SONAR models from \texttt{fairseq2}\footnote{\url{https://github.com/facebookresearch/SONAR}}.
Like Seamless, 
all encoders have $24$ layers and a feature dimension of $1024$. 
We extract representations from layers $\{1, 2, 4, 6, 8, 10, 12, 14, 16, 18, 20, 22, 23, 24\}$, the input embeddings, and the final speech and text embeddings after pooling, which are all of the same feature dimension of $1024$.

\subsection{SALMONN}
\textbf{S}peech \textbf{A}udio \textbf{L}anguage \textbf{M}usic \textbf{O}pen \textbf{N}eural \textbf{N}etwork (SALMONN; \citeauthor{salmonn}, \citeyear{salmonn}) is based on Vicuna\footnote{\url{https://huggingface.co/lmsys/vicuna-7b-v1.5}} \cite{vicuna2023}, a text-based LLM fine-tuned from Llama2 \cite{llama2} to follow text instructions.
Vicuna is finetuned with low-rank adaptation (LoRA) \cite{lora} to ingest inputs from audio features from the Whisper \cite{whisper} encoder and the BEATs \cite{beats} encoder.
Window-level Q-Former \cite{blip2,salmonn} is used to downsample the audio features with a window size of 0.33 second. 

Since SALMONN only accepts audio and text inputs simultaneously and the auditory and textual embeddings are given to Vicuna as one concatenated input, the extracted raw representations equal the concatenated speech and text representations. 
To analyze hidden speech and text representations separately, the raw representations are split into speech and text representations with the input length dimension.

We use the 7B version of SALMONN\footnote{\url{https://huggingface.co/tsinghua-ee/SALMONN-7B}}.
The decoder has 32 layers. 
We analyze layers $\{1, 2, 4, 6, 8, 10, 12, 14, 16, 18, 20, 22, 24, 26, 28,$
$30, 31, 32\}$, the speech encoder outputs before the Q-Former, and the textual and the auditory embeddings after the Q-Former. 
The speech encoder outputs before the Q-former has the feature size of $2048$, while the text embeddings, speech encoder outputs after the Q-former and all decoder layers have a feature size of $4096$. 
The different feature sizes cause no problem for SVCCA, as it can handle different input feature dimensions.

\section{Details on Selected Languages}
\label{appendix:language_details}

\begin{table*}[ht!]
    \centering
    \small
    \setlength\tabcolsep{3.5pt}
    \begin{tabular}{ccccccccccccccccccccccccccccccc}
    \toprule
    \textbf{Code} & 
    \textbf{Name} & 
    \textbf{Script} & 
    \textbf{Family} & 
    \makecell{\textbf{Resource} \\ \textbf{level}} & 
    \makecell{\textbf{Seamless \&} \\ \textbf{SALMONN}} & 
    \textbf{SONAR} & 
    \makecell{\textbf{\# sentences} \\ \textbf{original}} & 
    \textbf{Deduplicated} \\ 
    \midrule
    amh & Amharic & Ethiopic & Afro-Asiatic & low & \checkmark &  & 516 & 296 \\ 
    arb & Arabic & Arabic & Afro-Asiatic & high & \checkmark & \checkmark & 428 & 283 \\ 
    asm & Assamese & Bengali & Indo-European & low & & \checkmark & 984 & 349 \\ 
    bul & Bulgarian & Cyrillic & Indo-European & low & \checkmark & \checkmark & 658 & 344 \\ 
    cat & Catalan & Latin & Indo-European & high & \checkmark & \checkmark & 940 & 350 \\ 
    cmn & Chinese Mandarin & Hant & Sino-Tibetan & high & \checkmark & \checkmark & 945 & 349 \\ 
    deu & German & Latin & Indo-European & high & \checkmark & \checkmark & 862 & 347 \\
    ell & Greek & Greek & Indo-European & medium & \checkmark &  & 650 & 333 \\  
    eng & English & Latin & Indo-European & high & \checkmark & \checkmark & 647 & 350 \\ 
    est & Estonian & Latin & Uralic & medium & \checkmark & \checkmark & 893 & 345 \\ 
    fin & Finnish & Latin & Uralic & high & \checkmark & \checkmark & 918 & 348 \\ 
    fra & French & Latin & Indo-European & high & \checkmark & \checkmark & 676 & 332 \\ 
    heb & Hebrew & Hebrew & Afro-Asiatic & low & & \checkmark & 792 & 347 \\ 
    hin & Hindi & Devanagari & Indo-European & medium & \checkmark & \checkmark & 418 & 265 \\ 
    hye & Armenian & Armenic & Indo-European & low & \checkmark &  & 932 & 350 \\ 
    ind & Indonesian & Latin & Austronesian & medium & \checkmark & \checkmark & 687 & 328 \\ 
    ita & Italian & Latin & Indo-European & high & \checkmark & \checkmark & 865 & 346 \\ 
    jpn & Japanese & Japanese & Japonic & high & \checkmark & \checkmark & 650 & 321 \\          
    kat & Georgian & Georgian & Kartvelian & low & \checkmark &  & 979 & 350 \\ 
    khm & Khmer & Khmer & Austroasiatic & low & \checkmark &  & 949 & 335 \\ 
    kor & Korean & Korean & Koreanic & medium & \checkmark & \checkmark & 382 & 270 \\ 
    lao & Lao & Lao & Tai-Kadai & low & & \checkmark  & 405 & 260 \\ 
    lit & Lithuanian & Latin & Indo-European & low & \checkmark & \checkmark & 986 & 349 \\ 
    mal & Malayalam & Malayalam & Dravidian & low & & \checkmark  & 985 & 344 \\ 
    mar & Marathi & Devanagari & Indo-European & low & \checkmark & \checkmark & 1020 & 349 \\ 
    nld & Dutch & Latin & Indo-European & high & \checkmark & \checkmark & 364 & 251 \\ 
    pes & Persian & Arabic & Indo-European & low & \checkmark & \checkmark & 871 & 324 \\ 
    rus & Russian & Cryrillic & Indo-European & medium & \checkmark & \checkmark & 775 & 344 \\ 
    sna & Shona & Latin & Atlantic-Congo & low & \checkmark &  & 925 & 348 \\ 
    snd & Sindhi & Arabic & Indo-European & low & \checkmark & \checkmark & 980 & 350 \\ 
    swh & Swahili & Latin & Atlantic-Congo & low & & \checkmark  & 487 & 312 \\ 
    tam & Tamil & Tamil & Dravidian & medium & \checkmark & \checkmark & 591 & 336 \\ 
    tel & Telugu & Telugu & Dravidian & medium & & \checkmark  & 472 & 302 \\ 
    tha & Thai & Thai & Tai-Kadai & medium & \checkmark & \checkmark & 1020 & 349 \\ 
    tur & Turkish & Latin & Turkic & medium & \checkmark & \checkmark & 743 & 329 \\ 
    yue & Cantonese & Hant & Sino-Tibetan & low & \checkmark & \checkmark & 819 & 339 \\ 
    \bottomrule
    \end{tabular}
    \caption{List of analyzed languages.
    The column ``\# sentences original'' lists the number of sentences of a language from FLEURS. 
    The number of unique sentences in the test split for each language is given under the ``Deduplicated'' column.}
    \label{tab:language_statistics}
\end{table*}

We use the FLEURS \cite{fleurs} test split to extract hidden representations, and use the normalized transcriptions instead of raw transcriptions for the text representations.
As FLEURS has multiple utterances for the same sentence with different speakers, we remove these duplicates randomly, so that the same sentence only appears once in the audio set.
We choose 30 language from the 102 languages supported by FLEURS for analyses. 
While deciding on the languages, we maintain an even distribution of different language characteristics such as script, family and resource-level (high, medium and low as classified by \citet{seamless}). 
The statistics per language are in \autoref{tab:language_statistics}.
We analyze the the same set of languages for SeamlessM4T and SALMONN. 
For SONAR, as it uses monolingual audio encoders and does not cover support six languages from this set,
we replace them with other languages at the same resource-level.
A fully overlapping set of languages would have provided a cleaner experimental setup. 
However, since our conclusions are based on comparisons of similarity scores within the same model between modalities and languages, rather than across different models, we believe that the differing sets of languages do not compromise the validity of our findings.

For the cross-modal analysis (\S\ref{subsec:crossmodal}), each representation sets are reduced to the
first 251 representations, as this is the smallest number of input data without duplicates. 
For the cross-lingual analysis (\S\ref{subsec:crosslingual}), each intersects were
reduced to the first 194 intersecting representations for SeamlessM4T and SALMONN, and 192 for SONAR.

\end{document}